\documentclass[10pt,twocolumn,letterpaper]{article}

\usepackage{cvpr}              

\usepackage{bbm}

\definecolor{cvprblue}{rgb}{0.21,0.49,0.74}
\definecolor{lightblue}{rgb}{0.84,0.91,0.97}
\definecolor{skyblue}{rgb}{0.0, 0.75, 1.0}
\usepackage[pagebackref,breaklinks,colorlinks,allcolors=cvprblue]{hyperref}
\usepackage{adjustbox}
\usepackage[table]{xcolor}
\usepackage{amsmath}
\usepackage{multirow}
\newcommand{\q}[1]{`#1'}

\def\paperID{46156} 
\def\confName{CVPR}
\def\confYear{2026}

\title{\textsc{DocPrune}: Efficient Document Question Answering via  \\
            Background, Question, and Comprehension-aware Token Pruning}

\author{
Joonmyung Choi\textsuperscript{\rm 1}\hspace{0.2cm}
Sanghyeok Lee\textsuperscript{\rm 2}\hspace{0.2cm}
Jongha Kim\textsuperscript{\rm 1}\hspace{0.2cm}
Sehyung Kim\textsuperscript{\rm 1}\hspace{0.2cm}
Dohwan Ko\textsuperscript{\rm 1}\hspace{0.2cm} \\
Jihyung Kil\textsuperscript{\rm 3}\hspace{0.2cm}
Hyunwoo J. Kim\textsuperscript{\rm 2}\thanks{Corresponding author.}\hspace{0.2cm} \\
\textsuperscript{\rm 1}Korea University\hspace{0.8cm}
\textsuperscript{\rm 2}KAIST\hspace{0.8cm} 
\textsuperscript{\rm 3}Adobe Research\hspace{0.3cm} \\
\tt\small \{pizard, jonghakim, skim129, ikodoh\}@korea.ac.kr \\
\tt\small jkil@adobe.com \hspace{0.4cm} 
\tt\small \{sanghyeoklee, hyunwoojkim\}@kaist.ac.kr \\
}

\begin{document}
\maketitle
\begin{abstract}
Recent advances in vision–language models have demonstrated remarkable performance across diverse multi-modal tasks, including document question answering that leverages structured visual cues from text, tables, and figures.
However, unlike natural images, document images contain large backgrounds and only sparse supporting evidence, leading to the inefficient consumption of substantial computational resources, especially for long documents.
We observe that existing token-reduction methods for natural images and videos fall short in utilizing the structural sparsity unique to documents.
To address this, we propose \textsc{DocPrune}, a training-free and progressive document token pruning framework designed for efficient long-document understanding. 
The proposed method preserves only the essential tokens for the task while removing unnecessary ones, such as background or question-irrelevant tokens.
Moreover, it automatically selects the appropriate layers to initiate token pruning based on the model's level of comprehension.
Our experiments on the M3DocRAG show that \textsc{DocPrune} improves throughput by 3.0$\times$ and 3.3$\times$ in the encoder and decoder, respectively, while boosting the F1 score by +1.0, achieving both higher accuracy and efficiency without any additional training.
\end{abstract}    
\begin{figure}[t]
    \centering
    \includegraphics[trim=0 0 0 0,clip,width=0.42\textwidth]{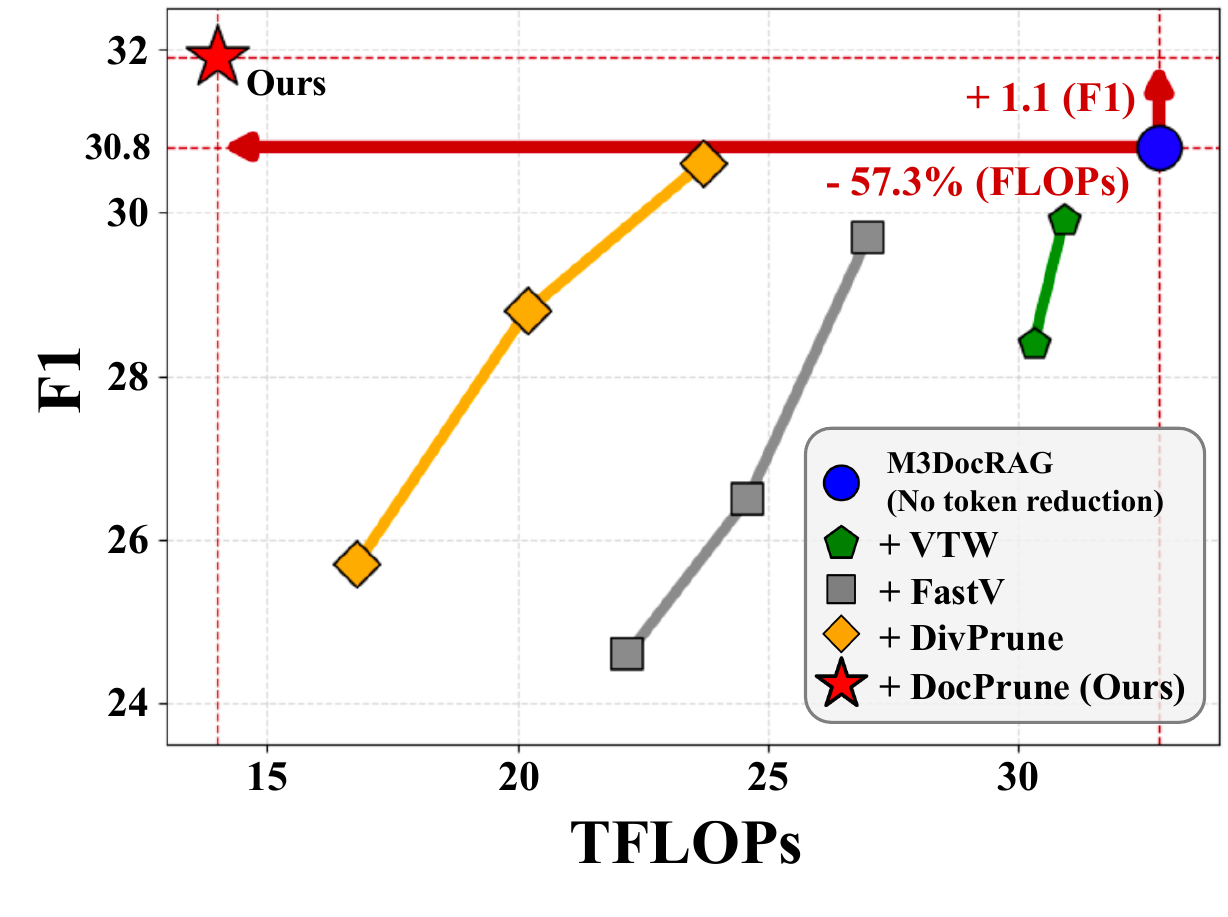}
        \vspace{-10pt}
        \caption{\textbf{Comparison of the token reduction methods.} Total TFLOPs of the encoder and decoder during QA are shown on the x-axis, and F1 scores on the y-axis. 
        Our DocPrune (`$\textcolor{red}{\bigstar}$') applied to the base model M3DocRAG (`\textcolor{blue}{$\bullet$}'), achieves the highest performance and the greatest complexity reduction compared to previous token reduction methods, even \emph{without} any additional training.} 
    \label{fig:fig_1}
\vspace{-15pt}
\end{figure}
\vspace{-10pt}
\section{Introduction}
Recent multimodal large language models (MLLMs)~\cite{wang2024qwen2,gemini,liu2023vit} have shown promising performance in document understanding tasks, including visual document question answering~\cite{mathew2021docvqa, mishra2019ocrvqa, huang2022layoutlmv3, han2025mdocagent, ma2024mmlongbench}, visual text grounding~\cite{Zhou_2025_ICCV, TGDoc, li2025towards}, and retrieval-augmented reasoning~\cite{cho2024m3docrag, tanaka2025vdocrag, mathur2024doc, yu2024visrag}.
However, document understanding remains computationally expensive, primarily due to the inherent sparsity of document layouts. They are often lengthy, containing hundreds of pages, but their content (\eg, text, tables, figures) is laid out on large backgrounds, resulting in thousands of visual tokens even on a single page. This substantially increases the computational cost for transformer-based models when understanding documents.

\begin{figure*}[t!]
  \centering
  \begin{minipage}[b]{0.32\textwidth}
      \centering
      \includegraphics[width=1.0\textwidth]{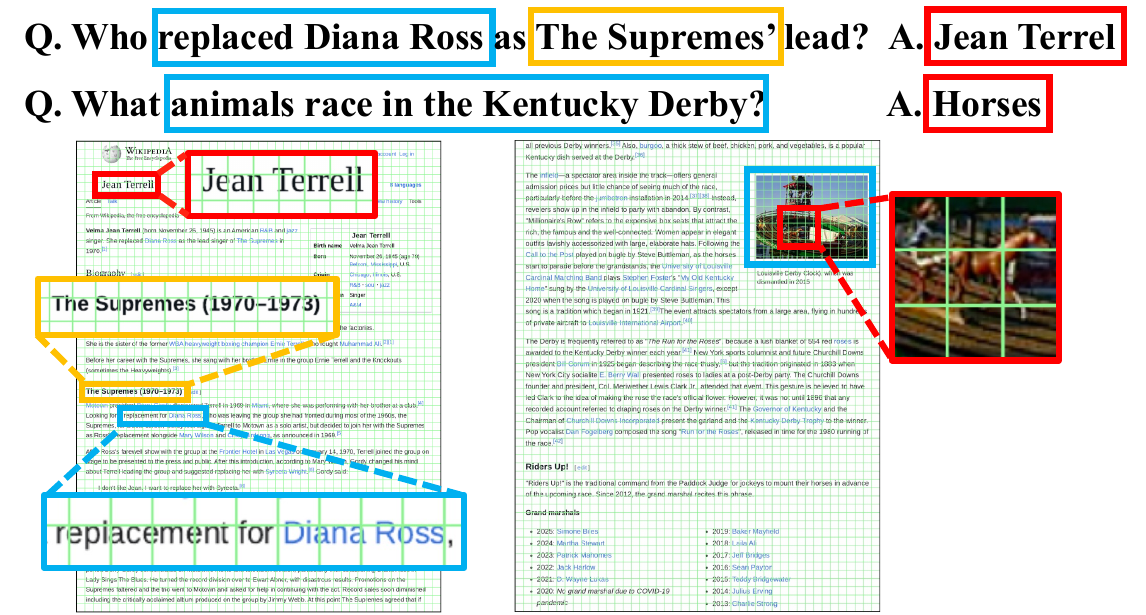}
      \caption{\textbf{Document layouts.} Question-relevant content regions occupy only small localized areas.}
      \label{fig:fig_2}
      \vspace{-5pt}
  \end{minipage}
  \hfill
  \begin{minipage}[b]{0.32\textwidth}
      \centering
      \includegraphics[width=0.95\textwidth]{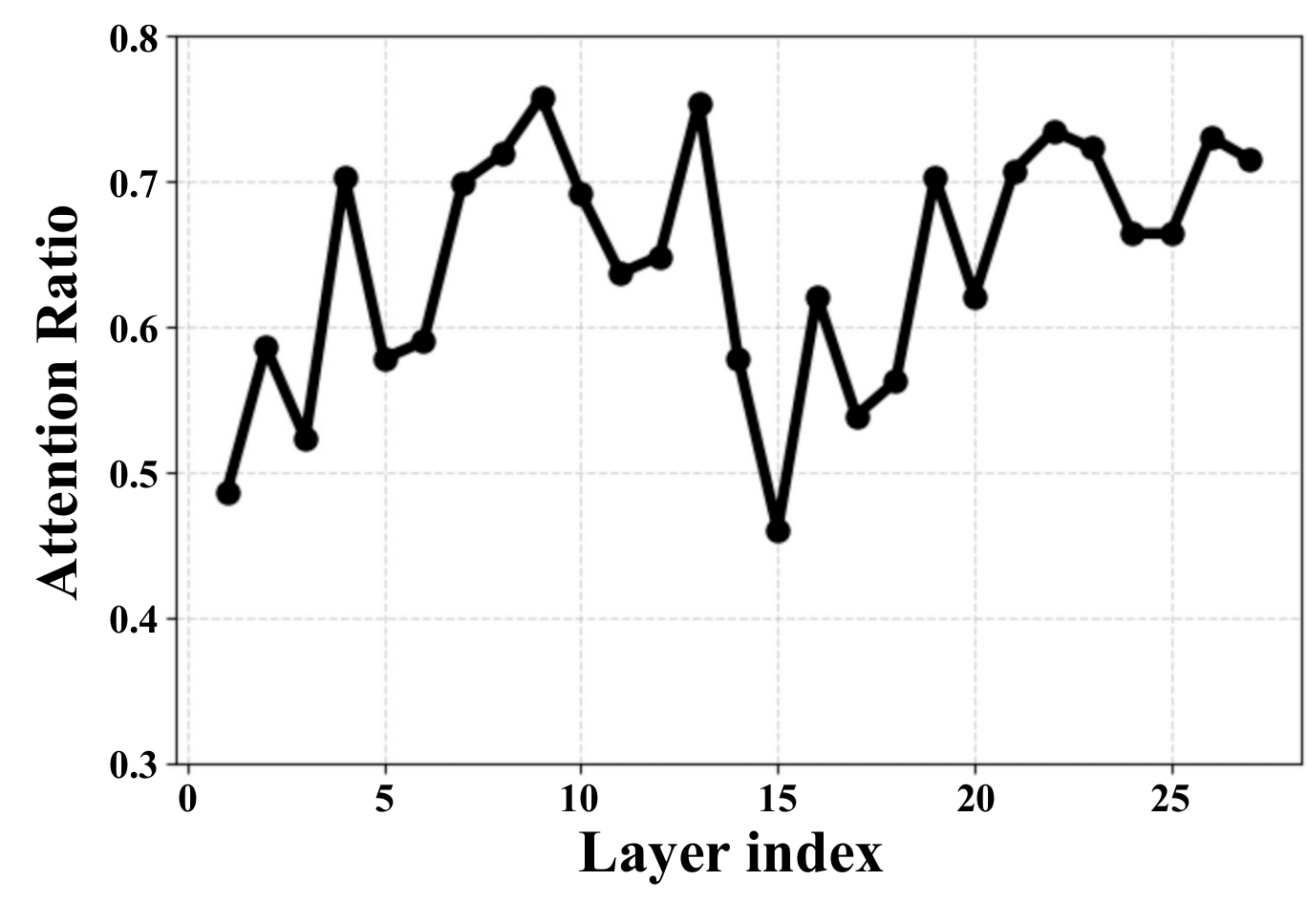}
      \vspace{-10pt}
      \caption{\textbf{Attention concentrated on the top 10\% most-attended tokens.}}
    \label{fig:fig_3}
  \end{minipage}
  \hfill
    \begin{minipage}[b]{0.32\textwidth}
    \centering
    \setlength{\tabcolsep}{2pt}
    \begin{tabular}{c|cc|cc}
      \toprule
      {Method} & TFlops & img/s & EM & F1  \\
      \midrule
      100\%   & 1.8 & 2.6  & 26.5 & 30.8 \\ 
      Top-10\%    & 0.2 & 16.8 & 25.7 & 29.7 \\ 
      \bottomrule
    \end{tabular}
    \captionof{table}{\textbf{Performance with full vs. top 10\% most-attended tokens.} Using only the top 10\% tokens maintains comparable performances with substantially lower computation.} 
    \label{tab:tab_1}
  \end{minipage}

\vspace{-10pt}
\end{figure*}

One way to address this challenge is token reduction which has been proven to be effective for natural images~\cite{long2023beyond, liang2022not, chen2024fastv, bolya2023tome, rao2021dynamicvit, zhang2025sparsevlm, lee2024multi} or videos~\cite{fu2024framefusion, choi2024vid, pollard2025video, shen2024longvu, choi2025representation, huang2025prunevid}, where nearby patches often contain similar visual information.
However, since document layouts are more structured, \eg, text lines, tables, and figures follow strict spatial organization, even small changes induced by token reduction can easily break text continuity or destroy important layout cues.
As a result, the previous works relying on visual redundancy exhibit considerable degradation.
Moreover, existing pruning approaches usually determine pruning layers through fixed heuristics~\cite{kim2026tabflash, chen2024fastv, bolya2023tome, zhang2025sparsevlm, lee2024multi}, neglecting how comprehension evolves across layers.
Without considering the layer-wise progression, pruning may happen too early or inconsistently, resulting in unstable performance and limited efficiency gains.

To address these limitations, we investigated token reduction from a document-centric perspective and identify three key observations.
First, large backgrounds often occupy a substantial portion of the page layout despite containing little semantic content, unnecessarily increasing computational overhead.
Second, even without the backgrounds, only a small portion of the remaining content is relevant to answering question.
Finally, determining pruning layers based on the model's level of comprehension is essential for stable document understanding. 
These findings suggest that token pruning methods designed for documents should consider both the document structure and the model's internal state.

\begin{figure*}[ht!]
\begin{center}
\includegraphics[width=0.95\textwidth]{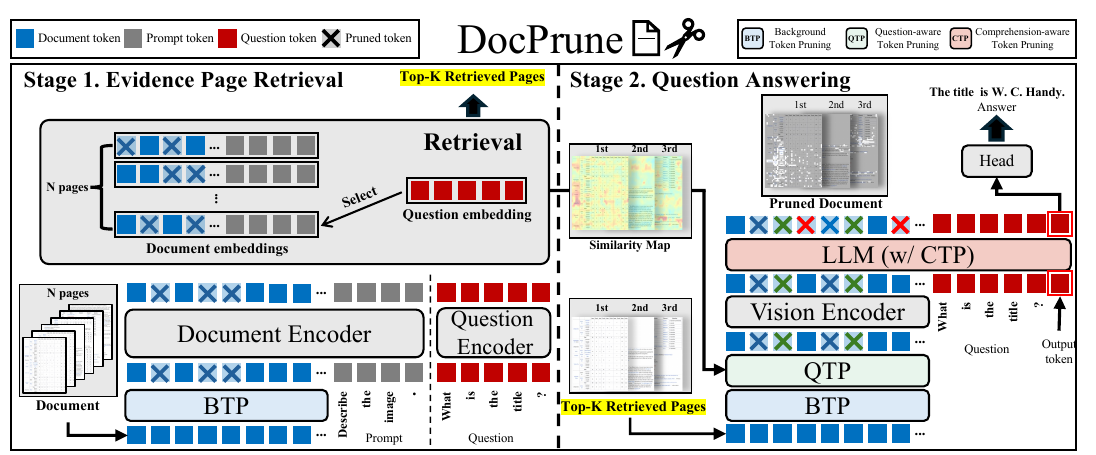}
\end{center}
\caption{ 
\textbf{Overview of DocPrune.}
DocPrune consists of two stages for document visual question answering. During the retrieval stage, Background Token Pruning (BTP) removes background tokens while keeping visual and textual content.
During the question answering stage, BTP and Question-aware Token Pruning (QTP) remove question-irrelevant tokens before the vision encoder input, and Comprehension-aware Token Pruning (CTP) further prunes tokens in the LLM decoder based on comprehension level for efficient inference.
}
\label{fig:main_figure}
\vspace{-0.5cm}
\end{figure*}

We thus propose \textsc{DocPrune} (\Cref{fig:main_figure}), a training-free framework for efficient long-document question answering.
Concretely, our system contains three stages. 
We first introduce Background Token Pruning (BTP), which removes non-informative background regions before encoding.
Then, Question-aware Token Pruning (QTP) further filters out unnecessary tokens using the similarity between question and document embeddings obtained from the retrieval stage.
Lastly, Comprehension-aware Token Pruning (CTP) monitors the layer-wise comprehension of the model through the L2 norm of the output token and triggers pruning once sufficient comprehension is achieved.
We observe that \textsc{DocPrune} improves encoder and decoder throughput by 3.0x and 3.3x while increasing F1 by 1.0 on M3DocRAG~\cite{cho2024m3docrag}.
These results demonstrate that leveraging both document structure and model comprehension enables efficient and accurate long-document QA \emph{without} any further training.
In summary, our contributions are threefold:
\begin{itemize}
    \item We first provide a comprehensive study of token redundancy and layer-wise comprehension in document understanding, revealing unique structural sparsity and comprehension patterns absent in natural images.
    \item We propose \textsc{DocPrune}, a training-free progressive token pruning framework for efficient long-document understanding.
    \item \textsc{DocPrune} achieves significant efficiency gains while maintaining or improving accuracy over the baseline.
\end{itemize}
\begin{figure*}[t!]
  \centering
  \begin{minipage}[b]{0.32\textwidth}
      \centering
      \includegraphics[width=0.99\textwidth]{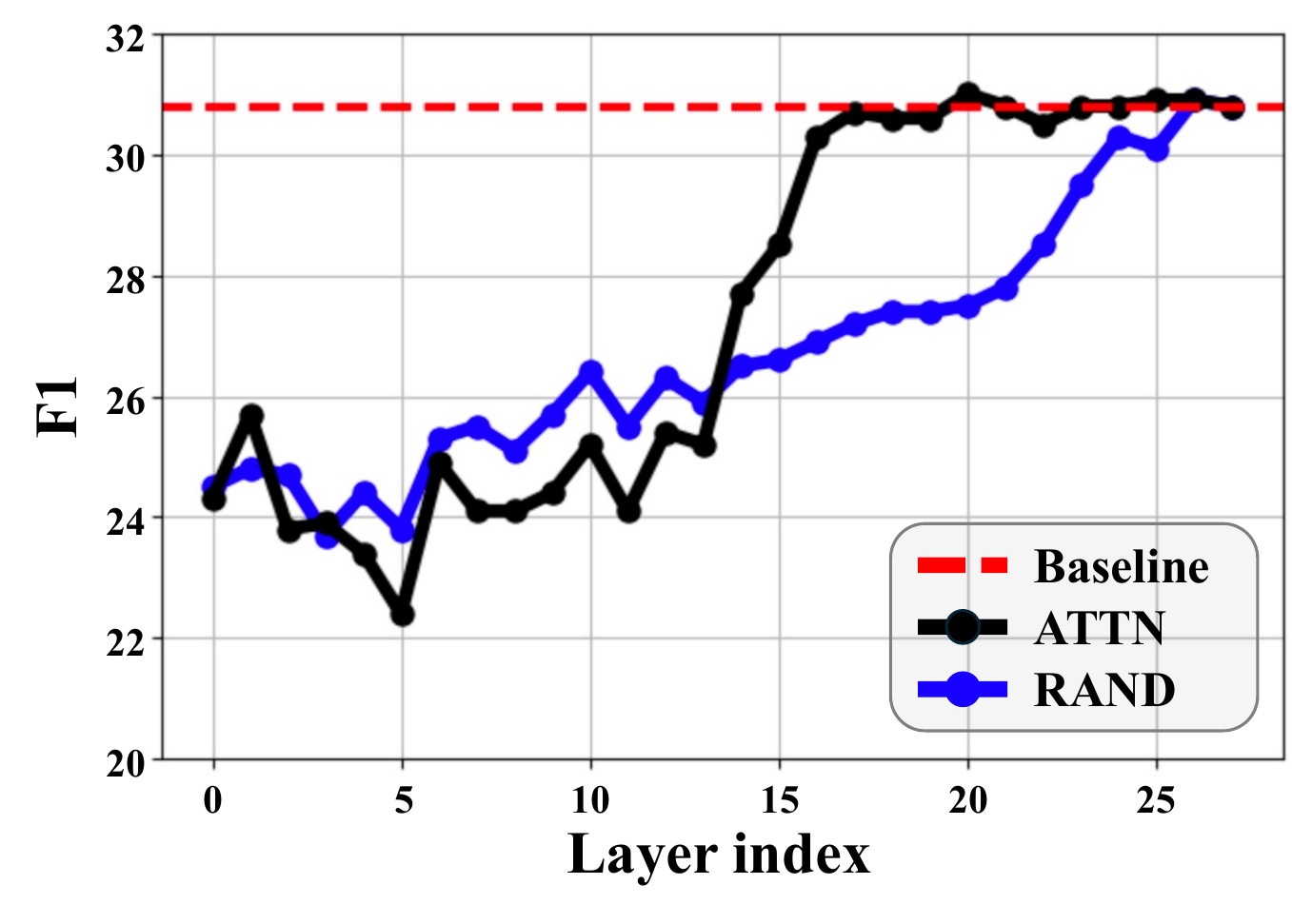}
      \vspace{-25pt}
      \caption{\textbf{Performance with 10\% tokens after pruning at each layer.}}
      \label{fig:fig_4}
  \end{minipage}
  \hfill
  \begin{minipage}[b]{0.32\textwidth}
      \centering
      \includegraphics[width=\textwidth]{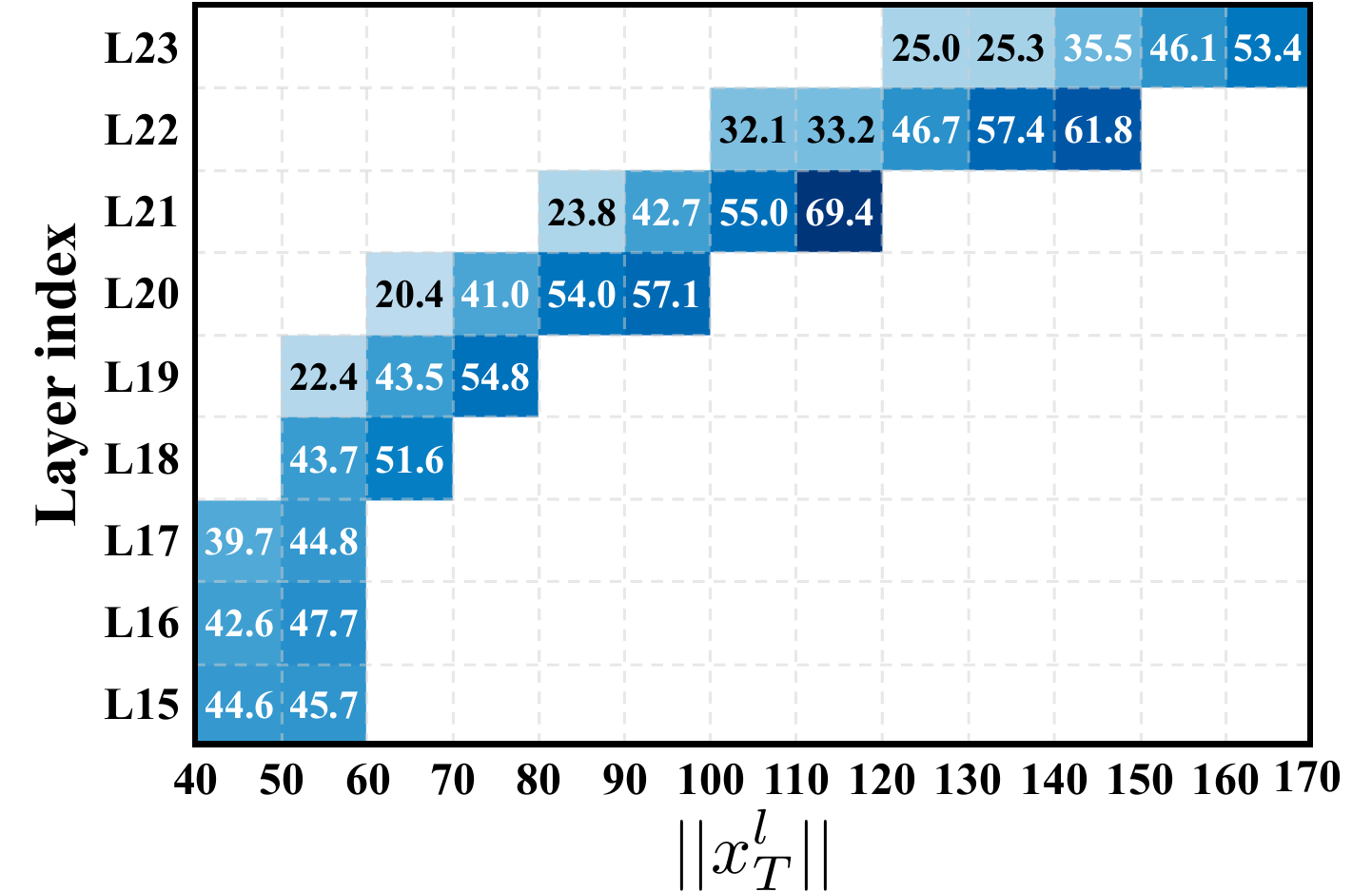}
      \vspace{-20pt}
      \caption{\textbf{Performance by layers and L2-norm.} Numbers in cells denote accuracy.}
    \label{fig:fig_5}
  \end{minipage}
  \hfill
  \begin{minipage}[b]{0.32\textwidth}
      \centering
      \includegraphics[width=0.9\textwidth]{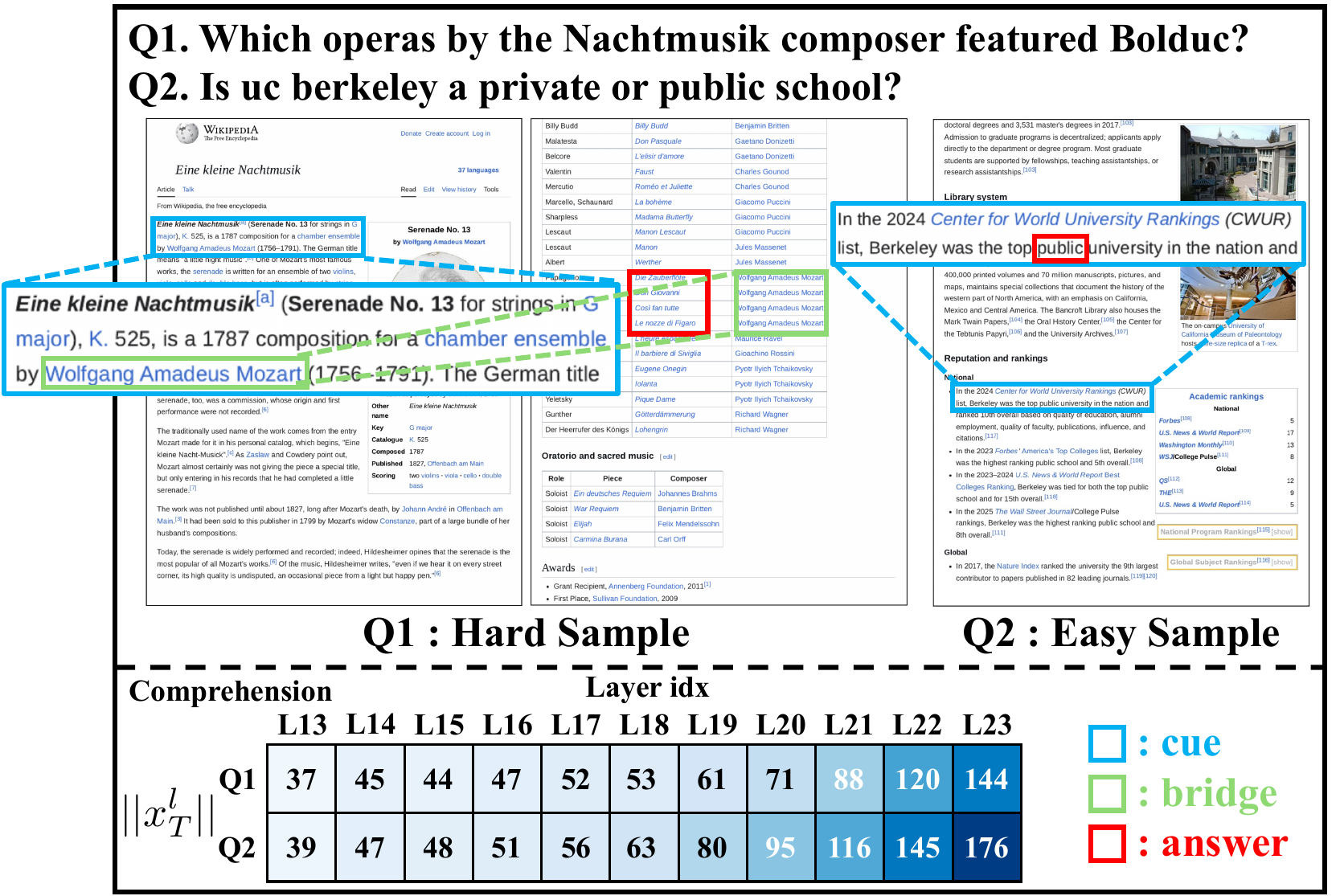}
      \vspace{-5pt}
      \caption{\textbf{Comprehension across hard vs. easy samples.}}
    \label{fig:fig_6}
  \end{minipage}
\vspace{-10pt}
\end{figure*}

\section{Key observations}
\label{sec:2}

To understand the unique challenges of document understanding, we begin with three key observations regarding background dominance of background, the distribution of question-relevant regions, and the effect of pruning across layers. These observations reveal why existing token-reduction methods are suboptimal and motivate our methods.

\noindent\textbf{Dominance of background regions.} 
Document images exhibit large non-informative background regions, such as page margins and inter-line spaces.
Although these regions carry little semantic information, they still incur substantial computational overhead.
To quantify this structural sparsity, we analyze document images from M3DocRAG.
For each document, the most frequent pixel value is defined as the background intensity, and the image is divided into non-overlapping $32\times 32$ patches as illustrated in~\Cref{fig:fig_2}.
A patch is labeled as background if all pixels show the same background value.
Our analysis shows that, on average, 36\% of patches correspond to background regions, indicating the prevalence of semantically redundant regions in document layouts.
This observation motivates Background Token Pruning (\Cref{sec:background_token_pruning}), which separates foreground content tokens from the background.

\noindent\textbf{Sparsity of question-relevant content.}
After background regions are excluded, we examine how much of the document content is actually required to answer the question.
As highlighted in~\Cref{fig:fig_2}, the supporting evidence for each answer is typically concentrated in a small, localized region within the document.
Consistent with this, attention analysis in~\Cref{fig:fig_3} reveals that the cumulative attention scores of the top 10\% most attentive tokens accounts for most of the attention (50\%$\sim$80\%).
In~\Cref{tab:tab_1}, we further examine the performance when retaining 10\% input tokens based on the attention from layer 20, considering that the cross-modal attention in deeper transformer layers best captures the semantic correspondence between the question and answer-relevant regions as discussed in~\cite{zhang2025flexselect}. 
Using only the top 10\% tokens for inference results in merely a 1.2-point drop in F1 score, while substantially decreasing the computational cost to $\times 1/9$ of the original FLOPs and achieving a 6.46$\times$ increase in throughput.
These results show that the model relies on a few informative tokens, whereas the remaining tokens merely increase computational overhead with minor contributions to the prediction.
Based on this, we propose Question-aware Token Pruning (\Cref{sec:question_aware_token_pruning}), which prunes question-irrelevant regions within the content area to preserve only tokens essential for answering.

\noindent\textbf{Layer-dependent pruning effects.}
Previous pruning methods selected pruning layers heuristically, which was effective in specific cases but lacked a generalizable principle for determining where pruning should occur.
To investigate this, we prune 90\% of tokens at each decoder layer under three settings: the full-token baseline (\textcolor{red}{red}), attention-guided pruning (black), and random pruning (\textcolor{blue}{blue}), as shown in~\Cref{fig:fig_4}.
Since the last token aggregates contextual information from the entire sequence during decoding, its attention distribution naturally reflects the contribution of visual tokens to the final answer.
In early layers, attention-guided pruning performs no better than random pruning, indicating that attention signals at shallow depths are not yet reliable for identifying salient tokens.
Beyond layer 15, its performance increases sharply and peaks around layer 20, suggesting that mid-to-late layers are effective for pruning with attention scores.
Although this analysis provides useful insight into where pruning is effective, it remains a post-hoc observation that lacks predictive capability, requires multiple runs, and offers no principled criterion for adaptively selecting pruning layers during inference.

To address this, we introduce a surrogate metric that predicts the level of the model's comprehension at each layer, defined as $c^l = ||x_T^l||$, where $x_T^l \in \mathbb{R}^d$ denotes the hidden state of the last token in the $l$-th layer.
\Cref{fig:fig_5} presents the relationship between $c^l$ and the sample-wise F1 scores on the M3DocRAG dataset, focusing on layers 15-23 where attention signals become reliably informative according to ~\Cref{fig:fig_4}.
Across all layers, samples with higher $c^l$ consistently exhibit better accuracies, and among samples within the same interval, those reaching such values at earlier layers tend to perform better.
Furthermore, easier samples tend to achieve higher $c^l$ values in earlier layers, whereas harder samples require deeper layers to reach comparable levels, as shown in~\Cref{fig:fig_6}.
These findings demonstrate that $c^l$ effectively captures the evolving confidence of the model, providing a reliable proxy for adaptively pruning tokens.
Based on this, we propose comprehension-aware token pruning, which adaptively drops the tokens once the model sufficiently comprehends the documents.

\section{Method}
\label{sec:3}
\subsection{Overall architecture}
\label{sec:document VLM}
We adopt a general retrieval-augmented document understanding pipeline, widely used in recent studies such as M3DocRAG~\cite{cho2024m3docrag}, SV-RAG~\cite{chen2024sv}, and VDocRAG~\cite{tanaka2025vdocrag}.

Given a question $q$ and a large collection of document pages $\mathcal{D}$, a retrieval model $f_\text{RET}$ retrieves a small set of top-$K$ relevant pages as:
\begin{equation}
    \tilde{\mathcal{D}}  =f_\text{RET}(q, \mathcal{D};K),
\end{equation}
where $\tilde{\mathcal{D}}$ contains the $K$ retrieved pages.
This step narrows the search space, limiting prediction to pages likely to contain supporting evidence. Subsequently, a multi-modal QA model $f_\text{QA}$ generates the final response $y$ using these retrieved pages:
\begin{equation}
    y=f_\text{QA}(q, \tilde{\mathcal{D}}).
\end{equation}
Built on this two-stage architecture, we propose \textsc{DocPrune}, a series of token pruning modules designed for both efficient and accurate multi-modal document understanding.
First, in~\Cref{sec:background_token_pruning}, Background Token Pruning (BTP) removes visually uninformative regions such as large backgrounds that dominate document layouts.
Next, in~\Cref{sec:question_aware_token_pruning}, Question-aware Token Pruning (QTP) filters the remaining tokens by assessing their semantic relevance to the query.
Finally, in~\Cref{sec:information_aware_token_pruning},  Comprehension-aware Token Pruning (CTP) adaptively prunes tokens during decoding based on the comprehension level of each layer, as approximated by the output-token representations.
This multi-stage design enables efficient processing across the two stages of retrieval and answer generation, forming the complete DocPrune framework illustrated in~\Cref{fig:main_figure}.

\subsection{Background Token Pruning}
\label{sec:background_token_pruning}
To efficiently reduce redundant visual tokens while retaining all content, we start with our first observation on ``Dominance of Background regions''.
In contrast to prior document VLMs that ignore the layout of documents, this stage explicitly detects and removes large background regions before the encoding stage begins.

Given an image $I\in \mathbb{R}^{H \times W \times 3}$, we first split the image into non-overlapping patches of size $P \times P$, resulting in the token set $T = \{t_i\in\mathbb{R}^{P\times P \times 3}\}^N_{i=1}$, where $N=\frac{W}{P} \times \frac{H}{P}$.
Each patch is converted into a grayscale map $\hat{t}_i \in \mathbb{R}^{P\times P}$ to simplify background detection.
We then estimate the dominant background intensity by computing its mode (most-frequent) value $m$ in the image, and measure the background ratio $R_i$  for each token as the proportion of pixels whose intensity is close to this background value:
\begin{equation}
R_i = \frac{1}{P^2} \sum_{p=1}^{P^2} \mathbbm{1} \left[\, |\hat{t}_i^{(p)} - m| < \tau_\text{e} \,\right],
\end{equation}
where $\hat{t}_i^{(p)}$ indicates the intensity of the $p$-th pixel of the $i$-th patch, $\tau_\text{e}$ denotes the error tolerance threshold for minor color variations in the background, and $\mathbbm{1}[ \cdot]$ is an indicator function.
Then, tokens with high background ratios are identified as background and discarded.
Consequently, the content tokens can be written as 
\begin{equation}
\tilde{T} = \{\, t_i \in T \mid R_i \le \tau_\text{bg} \,\},
\end{equation}
where $\tau_\text{bg}$ denotes the threshold that distinguishes background from content regions.
BTP is applied before the encoder of both page retrieval and question answering to eliminate redundant background regions, as illustrated in ~\Cref{fig:main_figure}.

\begin{figure*}[ht!]
\begin{center}
\includegraphics[width=0.95\textwidth]{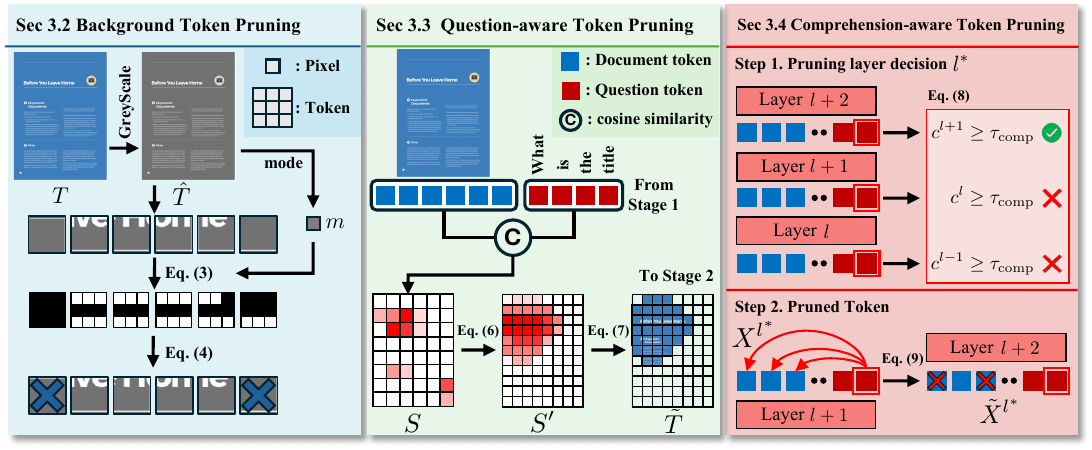}
\end{center}
\vspace{-15pt}
\caption{ 
\textbf{Illustration of proposed components.}
Background token pruning (BTP, Sec.\ref{sec:background_token_pruning}) removes background tokens while preserving content tokens.
Question-aware token pruning (QTP, Sec.\ref{sec:question_aware_token_pruning}) further eliminates content tokens that are highly irrelevant to the question based on retrieval scores.
Finally, Comprehension-aware token pruning (CTP, Sec.~\ref{sec:information_aware_token_pruning}) retains only a small subset of crucial tokens by considering both the model’s level of comprehension about the question and the attention based importance of each token.
}
\label{fig:method_figure}
\vspace{-10pt}
\end{figure*}
\begin{figure}[t]
    \centering
    \includegraphics[trim=0 0 0 0,clip,width=0.47\textwidth]{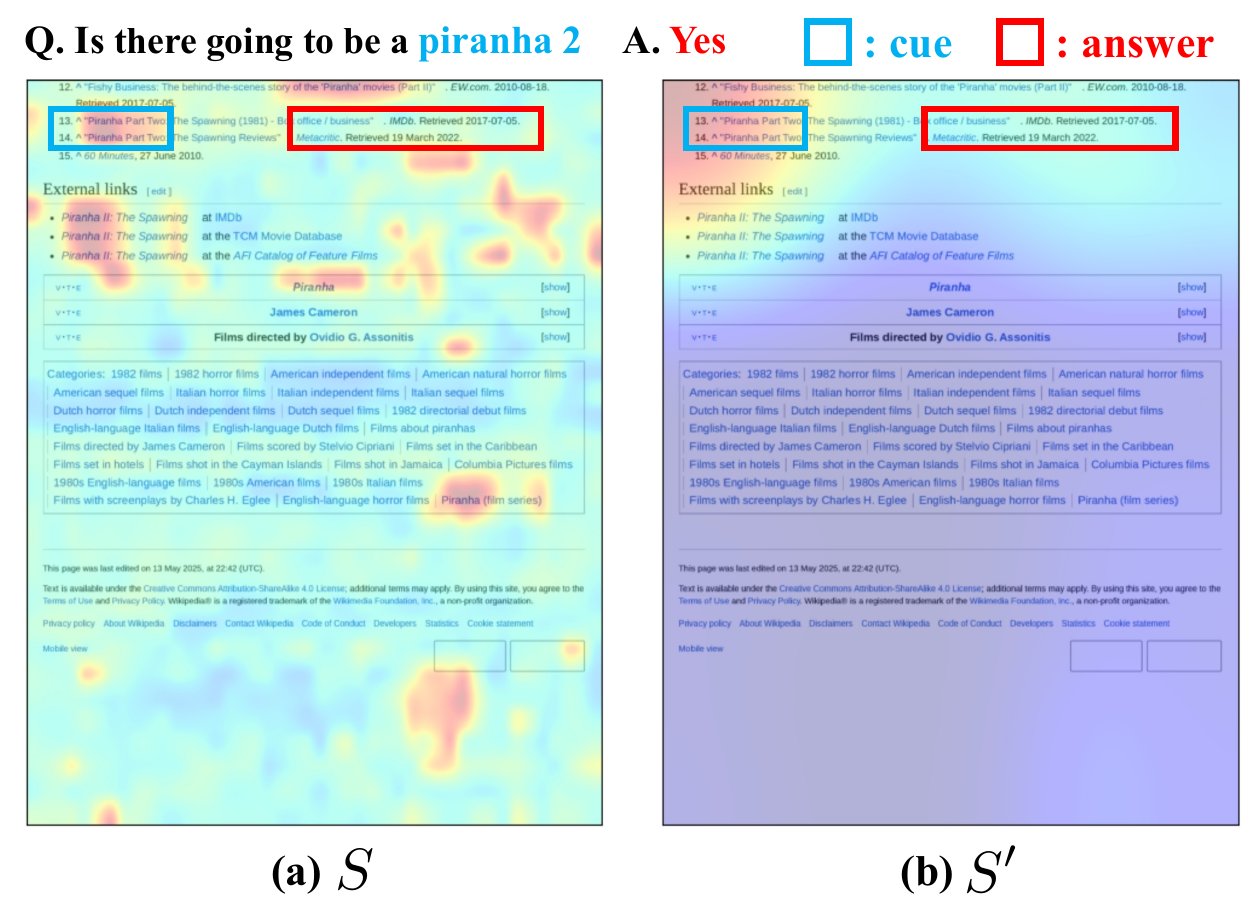}
\caption{ 
\textbf{Effect of Gaussian smoothing on the question–token similarity map.}
}
\vspace{-5pt}
\label{fig:fig_9}
\end{figure}

\subsection{Question-aware Token Pruning}
\label{sec:question_aware_token_pruning}
While BTP effectively eliminates visually non informative regions, it does not consider relevance to the given question.
As the second observation on ``Sparsity of question-relevant content'', we revealed that even among the content tokens, only a very small portion is truly relevant to the question.
To address this issue, we propose Question-aware Token Pruning (QTP) to retain only the question-relevant tokens.
To efficiently measure the relevance of the tokens in each document page, we leverage pre-computed document token embeddings $E^\text{doc}= \{\mathbf{e}^\text{doc}_i \in \mathbb{R}^{C} \}_{i=1}^{N_\text{doc}}$ and a question token embedding $E^\text{qst} = \{\mathbf{e}^\text{qst}_i\in \mathbb{R}^{C}\}^{N_\text{qst}}_{i=1}$ from the document retrieval stage, where $N_\text{doc}$ is the number of visual tokens per document and $C$ is the dimensionality. 
Then, we compute the cosine similarities between each document token and all question tokens, and aggregate them as
\begin{equation}
s_i = \sum_{j=1}^{N_\text{qst}} \cos(\mathbf{e}^\text{doc}_i, \mathbf{e}^\text{qst}_j),
\end{equation}
resulting in a set of relevance scores $S = \{ s_i \}_{i=1}^{N_\text{doc}}$.
When the retrieval and QA models use different encoders that operate at different feature-map resolutions (e.g., ColPali for retrieval followed by Qwen2-VL for QA), the resulting similarity map $S$ cannot be directly applied for token reduction in the QA encoder.
To bridge this resolution gap, we simply resize the similarity map $S$ to match the feature map resolution of the QA model through bilinear interpolation.
We empirically observe that this similarity map predominantly activates regions around question-related keywords, which are frequently located near the answers.
To broaden relevant regions and reduce localized noise, we apply Gaussian smoothing to the similarity map:
\begin{equation}
S' = G_{\sigma} * S,
\end{equation}
where $G_\sigma$ denotes a Gaussian kernel with standard deviation $\sigma$ and $*$ is the convolution operator, resulting in the smoothed relevance map $S'$ as shown in ~\Cref{fig:fig_9}.
Given the set of input tokens $T = \{ t_i \}_{i=1}^{N_\text{doc}}$, QTP outputs the tokens whose relevance scores exceed the threshold $\tau_\text{qst}$:
\begin{equation}
\tilde{T} = \{\, t_i \in T \mid S'_i \ge \tau_\text{qst} \,\}.
\end{equation}
The retained set $\tilde{T}$ is then passed into the QA vision encoder, as shown in~\Cref{fig:main_figure}.

\begin{table*}[t]
\centering
\small
\begin{adjustbox}{width=\textwidth}
\begin{tabular}{l|c|cccccc|ccccc|cc}
\toprule
\multirow{2}{*}{Method} & \multirow{2}{*}{\# Pages} & \multicolumn{2}{c}{Drop Rate} & \multicolumn{2}{c}{TFLOPs ($\downarrow$)} & \multicolumn{2}{c|}{Throughput} & \multicolumn{3}{c}{Evidence Modalities} & \multicolumn{2}{c}{Question Hops} & \multicolumn{2}{|c}{Overall} \\
\cmidrule(lr){3-4} \cmidrule(lr){5-6} \cmidrule(lr){7-8} \cmidrule(lr){9-11} \cmidrule(lr){12-13} \cmidrule(lr){14-15} 
& & ENC & DEC & ENC & DEC & ENC & DEC & Image & Table & Text & Single & Multi & EM & F1 \\
\midrule
Qwen2-VL (7B) & \multirow{5}{*}{1} & 0.00 & 0.00 & 14.82 & 18.02 & 2.3 & 2.6 & 26.6 & 25.6 & 37.2 & 35.8& 23.3 & 26.5& 30.8\\

+ FastV~\cite{chen2024fastv} \textcolor{gray}{$_{\text{ECCV2024}}$} &  & 0.00 & 0.50 & 14.82 & 9.84 & 2.3 & 4.6 & 24.3 & 22.9 & 35.5 & 32.8& 22.4 & 24.5& 28.6\\
+ DivPrune~\cite{alvar2025divprune} \textcolor{gray}{$_{\text{CVPR2025}}$} &  & 0.00 & 0.50 & 14.82 & 8.86 & 2.3 & 3.5 & 25.7 & 25.7 & 37.2 & 35.2& 23.8 & 26.4& 30.6\\
+ VTW~\cite{lin2025boosting} \textcolor{gray}{$_{\text{AAAI2025}}$} &  & 0.00 & \textbf{1.00} & 14.82 & 13.6 & 2.3 & 3.4 & 24.0 & 20.0 & 28.0 & 27.3& 20.0 & 21.0 & 24.3\\
\rowcolor{lightblue}
+ \textsc{DocPrune} (Ours) &  & \textbf{0.47} & 0.81 & \textbf{5.83} & \textbf{7.72} & \textbf{5.3} & \textbf{5.8} & \textbf{29.8} & \textbf{27.0} & \textbf{37.3} & \textbf{37.5} & \textbf{23.9} & \textbf{27.9} & \textbf{32.0}\\ \hline
Qwen2-VL (7B) & \multirow{5}{*}{2} & 0.00 & 0.00 & 29.64 & 38.24 & 1.2 & 1.3 & 31.5 & 28.8 & 39.9 & 40.8 & 24.2& 29.4 & 34.1\\
+ FastV~\cite{chen2024fastv} \textcolor{gray}{$_{\text{ECCV2024}}$} &  & 0.00 &0.50 & 29.64 & 20.19 & 1.2 & 2.4 & 26.0 & 25.8 & 37.9 & 36.3 & 23.2 & 26.4& 31.0\\
+ DivPrune~\cite{alvar2025divprune} \textcolor{gray}{$_{\text{CVPR2025}}$} &  & 0.00 & 0.50 & 29.64 & 18.02 & 1.2 & 1.5 & 30.6 & 27.4 & 39.2 & 39.2& 24.1 & 28.8& 33.1\\
+ VTW~\cite{lin2025boosting} \textcolor{gray}{$_{\text{AAAI2025}}$} &  & 0.00 & \textbf{1.00} & 29.64 & 28.76 & 1.2 & 1.8 & 28.7 & 20.2 & 25.6 & 27.0& 20.5 & 20.9& 24.4\\
\rowcolor{lightblue}
+ \textsc{DocPrune} (Ours) &  & \textbf{0.45} & 0.77 & \textbf{12.46} & \textbf{16.67} & \textbf{2.6} & \textbf{2.9} & \textbf{32.6} & \textbf{29.5} & \textbf{40.9} & \textbf{42.0} & \textbf{24.7} & \textbf{30.6} & \textbf{35.1}\\ \hline
Qwen2-VL (7B) & \multirow{5}{*}{4} & 0.00 & 0.00 & 59.28 & 86.27 & 0.6 & 0.6 & \textbf{33.7} & 29.5 & 43.2 & 43.9& \textbf{24.9} & 31.5 & 36.3\\
+ FastV~\cite{chen2024fastv} \textcolor{gray}{$_{\text{ECCV2024}}$} &  & 0.00 & 0.50 & 59.28 & 43.39 & 0.6 & 1.2 & 27.6 & 27.1 & 41.6 & 39.9& 23.8 & 28.7& 33.4\\
+ DivPrune~\cite{alvar2025divprune} \textcolor{gray}{$_{\text{CVPR2025}}$} &  & 0.00 & 0.50 & 59.28 & 38.25 & 0.6 & 0.4 & 33.0 & 28.4 & 42.6 & 43.4& 23.7 & 30.9& 35.5\\
+ VTW~\cite{lin2025boosting} \textcolor{gray}{$_{\text{AAAI2025}}$} &  & 0.00 & \textbf{1.00} & 59.28 & 64.79 & 0.6 & 0.8 & 28.2 & 19.3 & 27.3 & 27.7& 20.2 & 21.4& 24.7\\
\rowcolor{lightblue}
+ \textsc{DocPrune} (Ours) &  & \textbf{0.60} & 0.74 & \textbf{16.36} & \textbf{25.45} & \textbf{1.8} & \textbf{2.0} & 33.5 & \textbf{30.4} & \textbf{44.8} & \textbf{45.6} & 24.8 & \textbf{33.0} & \textbf{37.3}\\
\bottomrule
\end{tabular}
\end{adjustbox}
\caption{\textbf{Performance on M3DocRAG~\cite{cho2024m3docrag}.} ENC: vision encoder, DEC: LLM decoder, $\downarrow$: lower is better, throughput: samples/sec, EM: Exact Match. Better results within the same setting are marked \textbf{bold}. Accuracy by evidence modalities and question hops reported in F1.}
\label{tab:main}
\vspace{-5pt}
\end{table*}

\subsection{Comprehension-aware Token Pruning}
\label{sec:information_aware_token_pruning}
While BTP and QTP ensure that only question-relevant foreground tokens are retained, their propagation through the layers allows the model to gradually accumulate sufficient information for effective prediction.
As observed in our study of ``Layer-dependent pruning effects'', the L2-norm of the output token implicitly serves as a proxy for the model's comprehension at each layer.
Building on this insight, we propose Comprehension-aware Token Pruning (CTP) that adaptively determines the optimal timing and extent of pruning based on two criteria.

We first define a comprehension threshold $\tau_\text{comp}$ to determine the layer where pruning is applied.
At each decoder layer $l$, $X^l = \{x_i^l\}_{i=1}^N$ denotes the token representations, and $x_N^l \in \mathbb{R}^C$ is the last token at that layer.
We compute the L2-norm of the last token representation $c^l = ||x_N^l||$ to approximate the model's level of comprehension.
Once this value exceeds $\tau_\text{comp}$, the model is considered to have achieved sufficient comprehension, triggering the activation of pruning at that layer:
\begin{equation}
l^\ast = \min (\{l\;|\;c^l \geq \tau_\text{comp}\}).
\end{equation}
After the pruning layer $l^\ast$ is determined, tokens are pruned based on an attention threshold $\tau_\text{att}$.
The attention weights from the output token to all visual tokens are used as importance scores, denoted as $a^{l^\ast} \in \mathbb{R}^{N_\text{doc}}$, where $N_\text{doc}$ is the total number of visual tokens.
Lastly, tokens with attention values below $\tau_\text{att}$ are dropped before being propagated to the next layer:
\begin{equation}
    \tilde{X}^{l^\ast} = \{\, x_i^{l^\ast} \in X^{l^\ast} \mid a^{l^\ast}_i \ge \tau_\text{att} \}.
\end{equation}

%

\section{Experiments}
\label{sec:4}
\subsection{Implementation details}
We evaluate \textsc{DocPrune} across multiple models (M3DocRAG~\cite{cho2024m3docrag}, VDocRAG~\cite{tanaka2025vdocrag}) and diverse benchmarks (M3DocVQA, MMLongBench-Doc, ChartQA, SlideVQA, InfoVQA, DUDE).
We adopt M3DocRAG as our primary baseline, utilizing ColPali-v1~\cite{faysse2024colpali} for retrieval and Qwen2-VL (7B)~\cite{wang2024qwen2} for the QA model.
All pruning methods are applied to this baseline using their default configurations.
All experiments were conducted on a single NVIDIA RTX A6000 GPU with two AMD EPYC 7763 64-Core CPUs.

\subsection{Evaluation setup}
We follow the standard evaluation protocols for each benchmark.
For performance metrics, we report Accuracy (ACC), Exact Match (EM), and F1 scores averaged across the dataset are used for overall results. Also, F1 scores on the subset categorized by evidence modalities and question hops are reported.
To evaluate efficiency, we additionally measure the visual token drop rate (the proportion of tokens removed across the layers), TFLOPs, and throughput (samples/s) for both the vision encoder and LLM decoder, averaged across all samples.
These metrics are consistently applied across our primary baseline, M3DocRAG, and extended to the VDocRAG model.

\begin{figure*}[ht!]
\begin{center}
\includegraphics[width=0.95\textwidth]{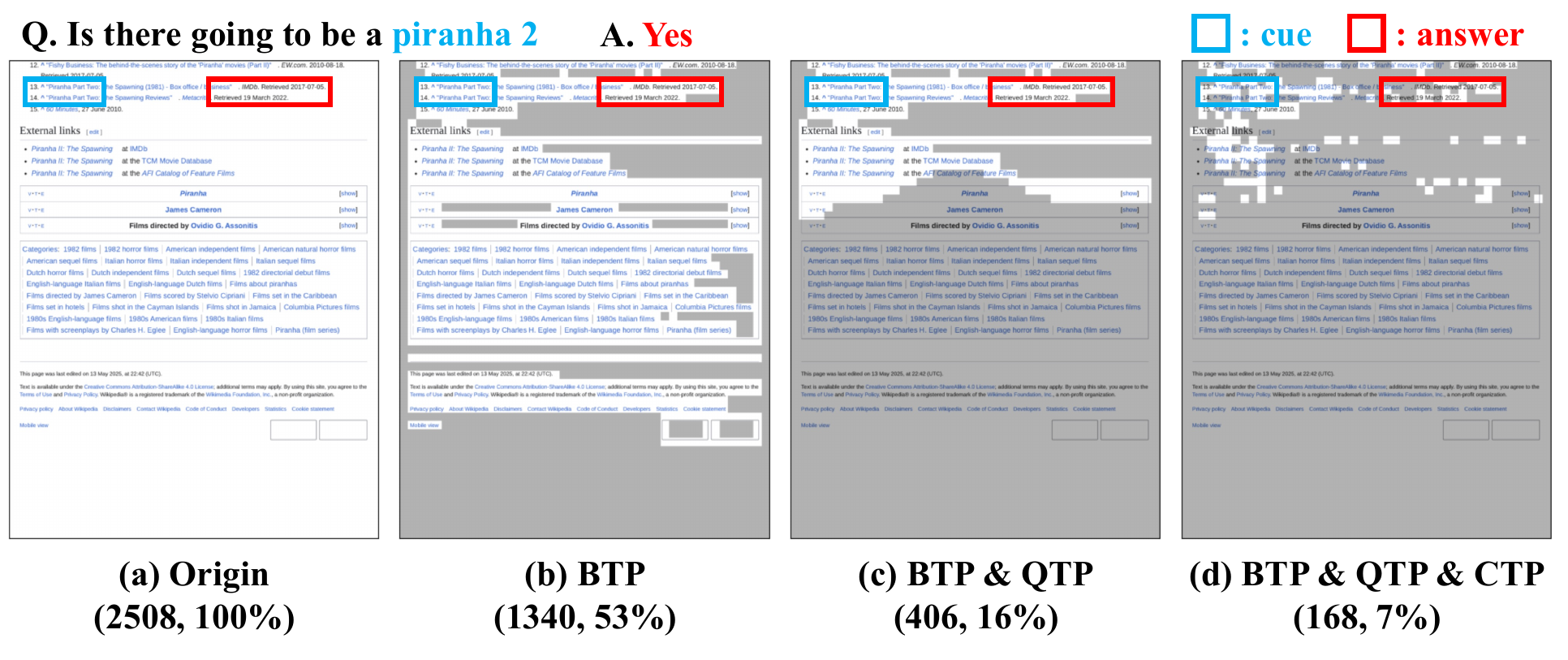}
\end{center}
\vspace{-20pt}
\caption{ 
\textbf{Qualitative results of DocPrune.}
(a) Original document image. 
(b) Background Token Pruning removes large background regions and preserves foreground content. 
(c) Question-aware Token Pruning retains tokens relevant to the question (cue, \textcolor{skyblue}{blue}). 
(d) Comprehension-aware Token Pruning further prunes redundant tokens based on layer-wise comprehension, leaving only answer-related tokens (answer, \textcolor{red}{red}).
The numbers in parentheses indicate the total number of remaining tokens at each stage and their relative percentage compared to the initial number of tokens.
}
\vspace{-10pt}
\label{fig:qual}
\end{figure*}

\begin{table}[t]
    \vspace{-5pt}
    \small
    \centering
    
    \begin{adjustbox}{width=\linewidth}
    \begin{tabular}{lc|cccc|cc}
    \toprule
    \multirow{2}{*}{Method} &  \multirow{2}{*}{\# Pages} & \multicolumn{2}{c}{TFLOPs($\downarrow$)} & \multicolumn{2}{c|}{Throughput} & \multicolumn{2}{c}{Overall} \\
    \cmidrule(lr){3-4} \cmidrule(lr){5-6} \cmidrule(lr){7-8} 
     & & ENC & DEC & ENC & DEC & ACC & F1 \\
    
    \hline
    Qwen2-VL (7B)  & \multirow{5}{*}{1} & 13.7 & 16.9 & 2.2 & 2.9 & 24.5 & 25.8 \\
    + \textsc{FastV}                &   & 13.7 & 10.1 & 2.5 & 4.5 & 20.4 & 21.1 \\
    + \textsc{DivPrune}             &   & 13.7 & 15.1 & 2.5 & 3.2 & 24.3 & 24.3 \\
    + \textsc{VTW}                  &   & 13.7 &  8.3 & 2.5 & 5.4 & 23.2 & 24.0 \\
    \rowcolor{lightblue}
    + \textsc{DocPrune (Ours)}      &   &  \textbf{5.2} &  \textbf{7.1} & \textbf{5.7} & \textbf{6.1} & \textbf{25.0} & \textbf{26.8} \\ 
    \hline
    Qwen2-VL (7B)  & \multirow{5}{*}{4} & 55.0 & 80.5 & 0.6 & 0.7 & 27.4 & 29.3 \\
    + \textsc{FastV}                &   & 55.0 & 40.6 & 0.6 & 1.3 & 22.8 & 24.4 \\
    + \textsc{DivPrune}             &   & 55.0 & 66.2 & 0.6 & 0.8 & 23.0 & 23.2 \\
    + \textsc{VTW}                  &   & 55.0 & 35.8 & 0.6 & 1.4 & 25.3 & 26.8 \\
    \rowcolor{lightblue}
    + \textsc{DocPrune (Ours)}      &   & \textbf{22.0} & \textbf{28.6} & \textbf{1.2} & \textbf{1.8} & \textbf{28.5} & \textbf{30.5} \\
    \hline
    Qwen2.5-VL (7B)  & & 13.7 & 16.9 & 3.9 & 2.9 & 27.3 & 27.9 \\
    \rowcolor{lightblue}
    + \textsc{DocPrune (Ours)}  & \multirow{-2}{*}{1} & \textbf{5.2} & \textbf{7.2} & \textbf{8.2} & \textbf{6.1} & \textbf{28.2} & \textbf{28.3} \\
    \hline
    Qwen2.5-VL (7B)  & & 55.0 & 80.5 & 1.0 & 0.7 & 31.5 & 33.4 \\
    \rowcolor{lightblue}
    + \textsc{DocPrune (Ours)}  & \multirow{-2}{*}{4} & \textbf{22.0} & \textbf{32.5} & \textbf{2.0} & \textbf{1.6} & \textbf{32.2} & \textbf{33.7} \\

    \bottomrule
    \end{tabular}
    \end{adjustbox}
    \vspace{-10pt}
    \caption{\textbf{Performance on MMLongBench-Doc.}}
    \vspace{-5pt}
    \label{tab:mmlongbench}
\end{table}
\begin{table}[t]
    \small
    \centering
    \begin{adjustbox}{width=\linewidth}
    \begin{tabular}{lc|cccc|cc}
    \toprule
    \multirow{2}{*}{Method} &  \multirow{2}{*}{Dataset} & \multicolumn{2}{c}{TFLOPS ($\downarrow$)} & \multicolumn{2}{c|}{Throughput} & \multicolumn{2}{c}{Performance} \\
    \cmidrule(lr){3-4} \cmidrule(lr){5-6} \cmidrule(lr){7-8} 
                     & & ENC & DEC & ENC & DEC & Single & All \\
    \midrule
    
    VDocRAG &  & 9.9 & 30.4 & 3.1 & 1.5 & 52.0 & \textbf{48.0} \\
    \rowcolor{lightblue}
    + \textsc{DocPrune (Ours)} & \multirow{-2}{*}{ChartQA} & \textbf{6.7} & \textbf{10.9} & \textbf{4.6} \scriptsize\textcolor{red}{($\times$1.5)} & \textbf{3.9} \scriptsize\textcolor{red}{($\times$2.6)} & \textbf{53.3} \scriptsize\textcolor{red}{(+1.3)} & \textbf{48.0} \\ 
    \hline
    VDocRAG &  & 9.9 & 30.2 & 3.1 & 1.6 & 44.2 & 42.0 \\
    \rowcolor{lightblue}
    + \textsc{DocPrune (Ours)} & \multirow{-2}{*}{SlideVQA} & \textbf{6.9} & \textbf{12.0} & \textbf{4.4} \scriptsize\textcolor{red}{($\times$1.4)} & \textbf{3.5} \scriptsize\textcolor{red}{($\times$2.2)} & \textbf{45.0} \scriptsize\textcolor{red}{(+0.8)} & \textbf{43.6} \scriptsize\textcolor{red}{(+1.6)} \\
    \hline
    VDocRAG &  & 9.9 & 33.0 & 3.1 & 1.4 & 56.2 & 49.2 \\
    \rowcolor{lightblue}
    + \textsc{DocPrune (Ours)} & \multirow{-2}{*}{InfoVQA} & \textbf{7.7} & \textbf{16.3} & \textbf{3.9} \scriptsize\textcolor{red}{($\times$1.3)} & \textbf{2.8} \scriptsize\textcolor{red}{($\times$2.0)} & 55.1 \scriptsize\textcolor{blue}{(-1.1)} & 49.8 \scriptsize\textcolor{red}{(+0.6)} \\   
    \hline
    VDocRAG &  & 9.9 & 34.9 & 3.2 & 1.4 & 48.5 & 44.0 \\
    \rowcolor{lightblue}
    + \textsc{DocPrune (Ours)} & \multirow{-2}{*}{DUDE} & \textbf{5.7} & \textbf{11.5} & \textbf{5.4} \scriptsize\textcolor{red}{($\times$1.7)} & \textbf{3.7} \scriptsize\textcolor{red}{($\times$2.6)} & \textbf{47.2} \scriptsize\textcolor{blue}{(-1.3)} & \textbf{43.8} \scriptsize\textcolor{blue}{(-0.2)} \\   
    
    \bottomrule
    \end{tabular}
    \end{adjustbox}
    \vspace{-10pt}
    \caption{\textbf{Performance on VDocRAG.} "Single" retrieve from dataset-specific pools; "All" uses a multi-domain pool.}
    \vspace{-10pt}
    \label{tab:vdocrag}
\end{table}

\subsection{Main results}
\noindent\textbf{Open-domain Doc-VQA.}
Tab.~\ref{tab:main} compares \textsc{DocPrune} against baseline and prior pruning methods (FastV~\cite{chen2024fastv}, DivPrune~\cite{alvar2025divprune}, VTW~\cite{lin2025boosting}) on the M3DocVQA benchmark using the M3DocRAG model.
Without additional training, \textsc{DocPrune} consistently improves both efficiency and accuracy. At top-4 retrieved pages, it reduces TFLOPs by over 70\% in both the visual encoder and LLM decoder, achieving up to 3.3$\times$ throughput gains while improving EM and F1 scores by 1.5 and 1.0 points, respectively.
In contrast, existing methods struggle with document-specific structures: VTW's layer-wise pruning harms fine-grained understanding, and DivPrune's iterative overhead even degrades throughput at high token counts.
\textsc{DocPrune} is the only approach that effectively reduces computation across the entire pipeline while surpassing the unpruned baseline.
To demonstrate generalization, we evaluated DocPrune additional model VDocRAG
(Tab.~\ref{tab:vdocrag}), DocPrune yields an average 2.4$\times$ speed up with competitive performance, proving its broad applicability to various DocQA tasks.

\noindent\textbf{Closed-domain Doc-VQA.}
Tab.~\ref{tab:mmlongbench} presents the performance of DocPrune on the closed-domain benchmark MMLongBench-Doc. The results consistently demonstrate that DocPrune outperforms all existing methods(FastV, VTW, and DivPrune), and even surpasses the unpruned baseline. Specifically, under the top-1 setting with Qwen2.5-VL model, OURS achieves throughpuit improvements of 2.1$\times$ in both the encoder and decoder compared to the baseline, while also yielding performance gains of +0.9 in ACC and +0.4 in F1.

\subsection{Analyses}
\label{sec:analysis}
Here, we analyze the behavior of \textsc{DocPrune}, including qualitative visualizations, component-wise ablations, and sensitivity studies of our pruning criteria and hyperparameters.
Hereafter, all experiments for analysis are conducted with the top-1 page setting unless specified.

\begin{table}[t]
\centering
\small
\begin{adjustbox}{width=\linewidth}
\begin{tabular}{cccccccc}
\toprule
\multirow{2}{*}{\# Pages} & \multirow{2}{*}{BTP} & \multirow{2}{*}{QTP} & \multirow{2}{*}{CTP} & \multicolumn{2}{c}{Throughput} & \multicolumn{2}{c}{Overall} \\
 \cmidrule(lr){5-6} \cmidrule(lr){7-8}
& & & & ENC & DEC & EM & F1 \\
\midrule
\multirow{4}{*}{1} & &  & & 2.3 & 2.6 & 26.5 & 30.8 \\
 &  \checkmark & & &  4.4 & 4.3 & 27.5 & 31.6 \\
 & \checkmark & \checkmark & & 5.2 & 5.0 & 27.7 & 31.8 \\
 &  \checkmark & \checkmark & \checkmark & 5.3 & 5.8 & 27.9 & 32.0 \\
\midrule
\multirow{4}{*}{2} & &  &  & 1.2 & 1.3 & 29.4 & 34.1\\
 & \checkmark & & & 2.0 & 2.1& 30.3 & 34.7\\
 & \checkmark & \checkmark& & 2.6 & 2.4& 30.4& 34.9\\
 & \checkmark & \checkmark & \checkmark& 2.6& 2.9 & 30.6& 35.1 \\
\midrule
\multirow{4}{*}{4} &  &  & & 0.6 & 0.6& 31.5 & 36.3\\
 &  \checkmark & & & 1.3 & 1.3 & 32.9 & 37.1\\
 &  \checkmark & \checkmark & & 1.8 & 1.7 & 32.6 & 36.9\\
 &  \checkmark & \checkmark & \checkmark & 1.8 & 2.0 & 33.0 & 37.3 \\
\midrule
\end{tabular}
\end{adjustbox}
\caption{\textbf{Ablation on components.} BTP: Background token pruning (Sec.~\ref{sec:background_token_pruning}), QTP: Question-aware token pruning (Sec.~\ref{sec:question_aware_token_pruning}), CTP: Comprehension-aware token pruning (Sec.~\ref{sec:information_aware_token_pruning}).
}
\label{tab:ablation_study}

\vspace{-10pt}

\end{table}

\noindent\textbf{Qualitative analysis.}
Fig.~\ref{fig:qual} presents qualitative examples illustrating how each stage of \textsc{DocPrune} progressively removes redundancy while preserving essential evidence.
In Fig.\ref{fig:qual}-(a), the original document image is encoded into 2,508 tokens, although only a small region is relevant to the cue or answer.
After applying BTP in Fig.\ref{fig:qual}-(b), background regions are removed while all content areas remain, reducing the tokens to 1,340.
QTP in Fig.\ref{fig:qual}-(c) further discards content irrelevant to the question, lowering the count to 460.
Finally, CTP in Fig.~\ref{fig:qual}-(d) retains only the most crucial tokens based on model comprehension, compressing the representation to 168 tokens, about 7 percent of the original, while still preserving all cue (\textcolor{blue}{blue}) and answer (\textcolor{red}{red}) regions.
These examples show that \textsc{DocPrune} removes substantial redundancy without harming the semantic evidence needed for accurate answers. 
We also observe cases where removing irrelevant regions improves predictions by guiding attention toward key evidence, with additional examples provided in the supplementary material. 
Overall, the progressive token reduction at each stage significantly improves end-to-end throughput while maintaining the information required for correct inference.

\noindent\textbf{Component ablations.}
Tab.~\ref{tab:ablation_study} presents the ablation results for three proposed components, BTP, QTP, and CTP.
As shown, both throughput and QA performance progressively are enhanced as each component is added, demonstrating the effectiveness of all proposed modules.
When all three methods are applied, the model achieves the best results, boosting throughput by 3$\times$ and 3.3$\times$ on the visual encoder and LLM decoder, respectively, while also improving EM and F1 scores by 1.5 and 1.0 points compared to the baseline under the top-4 page setting.

\noindent\textbf{Model comprehension criteria.}
Tab.~\ref{tab:information_criteria} compares various criteria for estimating the model's comprehension level to determine the optimal layer for CTP (Sec.~\ref{sec:information_aware_token_pruning}).
For all three methods, token pruning is performed only once at the identified layer.
\q{Entropy} initiates pruning when the output distribution's uncertainty is sufficiently low.
In contrast, both \q{Feature $\Delta$} which measures the representation difference between successive layers~\cite{choi2025representation}, and our proposed \q{L2 norm} trigger the drop once their values exceed a specific threshold, indicating sufficient information accumulation.
Although entropy and Feature $\Delta$ yield moderate improvements, the L2 norm achieves the best balance between throughput and accuracy. 
Further analysis in the supplementary material confirms that this metric is highly correlated with QA performance, serving as the most reliable proxy for quantifying model comprehension.
\section{Related Work}
\noindent\textbf{Document Understanding.}
Understanding visually rich documents requires reasoning over both textual and structural cues, often across multiple pages and modalities.
Among the various tasks in this domain, Document Visual Question Answering (DocVQA) focuses on answering questions grounded in both textual and visual information within documents.
Early work~\cite{mishra2019ocrvqa,mathew2021docvqa} relied on OCR-based, single-page pipelines that struggled with long contexts and complex layouts.
Recent studies introduce multi-page and multimodal benchmarks~\cite{ma2024mmlongbench} that highlight new challenges in retrieval and reasoning across extended document contexts.
Advances in Vision–Language Models (VLMs)~\cite{liu2023vit,li2024llamavid,li2025mini,chen2024internvl} have enabled OCR-free approaches that treat document pages as images, allowing end-to-end visual understanding.
ColPali~\cite{faysse2024colpali} extends this idea by encoding each page as a visual embedding for retrieval, while subsequent works such as VisRAG and VDocRAG~\cite{tanaka2025vdocrag} leverage page-level visual inputs to avoid the limitations of text parsing. M3DocRAG~\cite{cho2024m3docrag} further develops a multimodal retrieval-augmented generation (RAG) pipeline that retrieves relevant document pages and performs vision–language understanding, and MDocAgent~\cite{han2025mdocagent} extends this with multi-agent retrieval and cross-modal coordination.
Building on these trends, our work focuses on efficiency within this setting through lightweight, training-free token pruning guided by document content and question semantics.

\begin{table}[t]
\centering
\small
\begin{tabular}{cccccc}
\toprule
\multirow{2}{*}{Criteria} & \multicolumn{2}{c}{Throughput} & \multicolumn{2}{c}{Overall} \\
 \cmidrule(lr){2-3} \cmidrule(lr){4-5}
 & ENC & DEC & EM & F1 \\
\midrule
Baseline & \textbf{2.3} & 2.6 & 26.5 & 30.8 \\
Entropy & \textbf{2.3} & 2.9 & 26.3 & 30.2 \\
Feature $\Delta$ & \textbf{2.3} & \textbf{3.5} & 26.3 & 30.5 \\
\rowcolor{lightblue}
L2 Norm (Ours) & \textbf{2.3} & \textbf{3.5} & \textbf{26.8} & \textbf{30.9} \\
\bottomrule
\end{tabular}
\caption{\textbf{Analysis of comprehension criteria in CTP.} BTP and QTP are not applied to isolate the impact of the criteria on CTP.}
\vspace{-12pt}
\label{tab:information_criteria}
\end{table}

\noindent\textbf{Visual Token Pruning for VLMs.}
Vision–Language Models (VLMs) such as LLaVA~\cite{liu2023vit}, Video-LLaMA~\cite{zhang2023video} process far more visual tokens than language tokens, resulting in high inference costs on detailed or long inputs. To mitigate this, recent work has explored two main directions. LLaMA-VID~\cite{li2024llamavid} and DeCo~\cite{yao2024deco} compress visual features through projection or adaptive pooling, while methods such as FastV~\cite{chen2024fastv}, VoCo-LLaMA~\cite{ye2025voco}, and SparseVLM~\cite{zhang2025sparsevlm} prune low-importance tokens during decoding. However, existing methods mainly target images or videos, whereas our \textsc{DocPrune} focuses on documents by leveraging question and structural cues for token pruning.
\section{Conclusion}
In this work, we introduced DocPrune, a training-free progressive token pruning framework designed to improve the efficiency of document visual QA.
Unlike prior pruning methods primarily developed for images and videos with empirically chosen pruning layers, DocPrune exploits the structured layout of documents and employs a comprehension-aware criterion to select pruning layers adaptively.
Through three complementary stages, Background Token Pruning (BTP), which removes large non-informative background regions; Question-aware Token Pruning (QTP), which removes tokens showing weak alignment with the question, and Comprehension-aware Token Pruning (CTP), which adaptively prunes tokens based on the model's layer-wise comprehension, DocPrune progressively eliminates redundant tokens in a way that reflects the objective of each stage.
The experiments show that DocPrune consistently outperforms prior pruning methods even with lower computation, demonstrating its ability to achieve accurate and efficient document question answering.

\section*{Acknowledgements.}
This work was supported by the Institute of Information \& Communications Technology Planning \& Evaluation (IITP) grant funded by the Korea government (MSIT) (No. RS-2024-00443251, Accurate and Safe Multimodal, Multilingual Personalized AI Tutors, 40\%; No. RS-2024-00457882, AI Research Hub Project, 30\%), and by the InnoCORE program of the Ministry of Science and ICT (N10250156, 30\%).


{
    \small
    \bibliographystyle{ieeenat_fullname}
    \bibliography{main}
}

\clearpage
\maketitlesupplementary
\setcounter{section}{0}
\renewcommand{\thesection}{\Alph{section}}
\setcounter{table}{0}
\renewcommand{\thetable}{\Alph{table}}
\setcounter{figure}{0}
\renewcommand{\thefigure}{\Alph{figure}}

\renewcommand{\thetable}{\Alph{table}}
\renewcommand{\thefigure}{\Alph{figure}}

\section{Experimental settings}
In this section, we delineate implementation details for applying \textsc{DocPrune} to M3DocRAG~\cite{cho2024m3docrag} and outline the hyperparameter choices for both our method and the baseline token pruning approaches.

\noindent\textbf{Implementation details.}
All results reported in the main paper are obtained by evaluating the model in a training-free manner, without any additional training or fine-tuning.
When using Qwen2-VL for question answering in M3DocRAG~\cite{cho2024m3docrag}, a spatial token merger with a 2$\times$2 grid is inserted after the vision encoder.
Note that applying BTP or QTP with arbitrary token pruning in the encoder stage would break this grid structure and cause the merger to behave incorrectly.
To preserve compatibility, \textsc{DocPrune} applies BTP and QTP at the encoder input by grouping spatial tokens into 2$\times$2 blocks and pruning at the block level, so that the merger consistently receives a valid token layout.
We also adapt \textsc{DocPrune} to remain compatible with FlashAttention~\cite{dao2023flashattention}.
While FlashAttention reduces memory and computational overhead by avoiding storage of the full attention matrix in HBM, this makes token-wise attention scores unavailable to our CTP module.
To address this, we simply recompute attention only for the last token using a reduced query only at the selected layer, providing the attention scores for token pruning at negligible additional cost.

\begin{table}[t]
    \centering
    \begin{adjustbox}{width=\linewidth}
    \begin{tabular}{ccccc | ccccc}
        \toprule
        \multicolumn{1}{c}{\multirow{2}{*}[-3pt]{\textbf{Value}}} & \multicolumn{2}{c}{\textbf{Throughput}} & \multicolumn{2}{c|}{\textbf{Overall}} &
        \multicolumn{1}{c}{\multirow{2}{*}[-3pt]{\textbf{Value}}} & \multicolumn{2}{c}{\textbf{Throughput}} & \multicolumn{2}{c}{\textbf{Overall}}\\
        \cmidrule(lr){2-5} \cmidrule(lr){7-10}
        & ENC & DEC & EM & F1 & & ENC & DEC & EM & F1 \\
        \midrule
        \multicolumn{5}{c}{\textbf{(a) Background threshold $\tau_{\text{bg}}$}} & \multicolumn{5}{c}{\textbf{(b) Relevance threshold $\tau_{\text{qst}}$}}\\
        1.0 & 4.9 & 5.5 & 28.1 & 32.1 & 0.1 & 4.5 & 5.0 & 27.9 & 31.9\\
        \cellcolor{lightblue} 0.9 & \cellcolor{lightblue}5.3 & \cellcolor{lightblue}5.8 & \cellcolor{lightblue}27.9 & \cellcolor{lightblue}32.0 & 0.2 & 4.8 & 5.5 & 27.7 & 31.9\\
        0.8 & 6.1 & 6.4 & 27.3 & 31.3 & \cellcolor{lightblue}0.3 & \cellcolor{lightblue}5.3 & \cellcolor{lightblue}5.8 & \cellcolor{lightblue}27.9 & \cellcolor{lightblue}32.0\\
        0.7 & 7.9 & 7.5 & 26.9 & 30.9 & 0.4 & 5.8 & 6.3 & 27.4 & 31.2\\
        \midrule
        \multicolumn{5}{c}{\textbf{(c) Comprehension threshold $\tau_{\text{comp}}$}} & \multicolumn{5}{c}{\textbf{(d) Attention threshold $\tau_{\text{att}}$}}\\
        60 & 5.0 & 5.9 & 27.8 & 32.0 & 0.1 & 5.2 & 5.2 & 27.7 & 31.8\\
        \cellcolor{lightblue}65 & \cellcolor{lightblue}5.3 & \cellcolor{lightblue}5.8 & \cellcolor{lightblue}27.9 & \cellcolor{lightblue}32.0 & 0.3 & 5.3 & 5.7 & 27.7 & 31.8\\
        70 & 5.3 & 5.7 & 27.9 & 32.0 & \cellcolor{lightblue}0.5 & \cellcolor{lightblue}5.3 & \cellcolor{lightblue}5.8 & \cellcolor{lightblue}27.9 & \cellcolor{lightblue}32.0\\
        75 & 5.3 & 5.7 & 27.9 & 31.9 & 0.7 & 5.3 & 5.9 & 27.8 & 32.0\\
        \bottomrule
    \end{tabular}
    \end{adjustbox}
    \caption{\textbf{Sensitivity analysis on M3DocVQA.} The highlighted row indicates the default settings for \textsc{DocPrune}.}
    \vspace{-10pt}
    \label{tab:sensitivity_analysis}
\end{table}
\begin{table}[t]
\centering
\begin{tabular}{c|c|cccccc}
\toprule
\multirow{2}{*}{\textbf{Page}} 
& \textbf{RET} 
& \multicolumn{5}{c}{\textbf{QA}} \\
\cmidrule(lr){2-2} \cmidrule(lr){3-7}
& $\tau_{\text{bg}}$
& $\tau_{\text{bg}}$
& $\tau_{e}$
& $\tau_{q}$
& $\tau_{\text{info}}$
& $\tau_{\text{att}}$ \\
\midrule
1 & 0.9 & 0.9 & 1 & 0.3 & 65 & 0.5 \\
2 & 1.0 & 1.0 & 1 & 0.3 & 60 & 0.25 \\
4 & 1.0 & 0.8 & 1 & 0.4 & 45 & 0.075 \\
\bottomrule
\end{tabular}
\caption{\textbf{Hyperparameters for \textsc{M3DocVQA}.}}
\vspace{-15pt}
\label{tab:hyperparameter}
\end{table}

\noindent\textbf{Sensitivity analysis.}
In Tab.~\ref{tab:sensitivity_analysis}, we present a sensitivity analysis of the hyperparameters.
Notably, \textsc{DocPrune} consistently surpasses all other pruning methods in all metrics, regardless of hyperparameter settings, demonstrating its robustness and superiority.
In detail, adjusting $\tau_{\text{bg}}$ and $\tau_{\text{qst}}$ allows flexible control over the trade-off between throughput and QA accuracy (Tab.~\ref{tab:sensitivity_analysis}-(a) and (b)).
Moreover, varying $\tau_{\text{comp}}$ and $\tau_{\text{att}}$ results in negligible performance fluctuation (Tab.~\ref{tab:sensitivity_analysis}-(c) and (d)), demonstrating the robustness of the model.

\noindent\textbf{Hyperparameters for \textsc{DocPrune}.}
For hyperparameter choice, since the number of pages affects the distribution of visual features and attention, we use separate hyperparameter sets for 1, 2, and 4 page inputs.
Specifically, we perform a grid search over $\tau_\text{bg}$ and $\tau_q$ with a step size of 0.1, and over $\tau_\text{info}$ with a step size of 5.
For $\tau_\text{att}$, we adopt finer step sizes of [0.1, 0.05, 0.025] as pages increase, as attention scores become more dispersed when the number of visual tokens increases.
The final hyperparameters are summarized in Tab.~\ref{tab:hyperparameter}.

\noindent\textbf{Hyperparameters for previous pruning methods.} 
For a fair comparison with previous works, we tune FastV~\cite{chen2024fastv}, DivPrune~\cite{alvar2025divprune}, and VTW~\cite{lin2025boosting} using the search ranges specified in their original configurations.
For clarity, we here use the same notation used in previous works, where $K$ denotes the drop layer and $R$ denotes the pruning ratio.
FastV is originally evaluated with $l \in \{0, 2, 3, 5\}$ and $r \in \{0.5, 0.75, 0.9\}$ in their paper, and $l=2$, $r=0.5$ consistently perform best in our setting.
In DivPrune, the original pruning ratio of $r=0.902$ is heuristic and yields performance similar to random pruning in our document setting. Therefore, we additionally explore $r \in \{0.3, 0.5, 0.7, 0.9\}$, and report the result with $r=0.5$, which achieves the best results.
DivPrune does select a drop-layer $l$ because it prunes tokens immediately before the decoder.
For VTW, the results are reported with $l \in \{16, 20\}$ with an original pruning ratio of $r=1.0$, and we find that $l=20$ yields the best results across all page counts.

\section{Qualitative analysis of removing irrelevant regions for QA} 
\begin{figure*}[ht!]
\begin{center}
\includegraphics[width=0.80\textwidth]{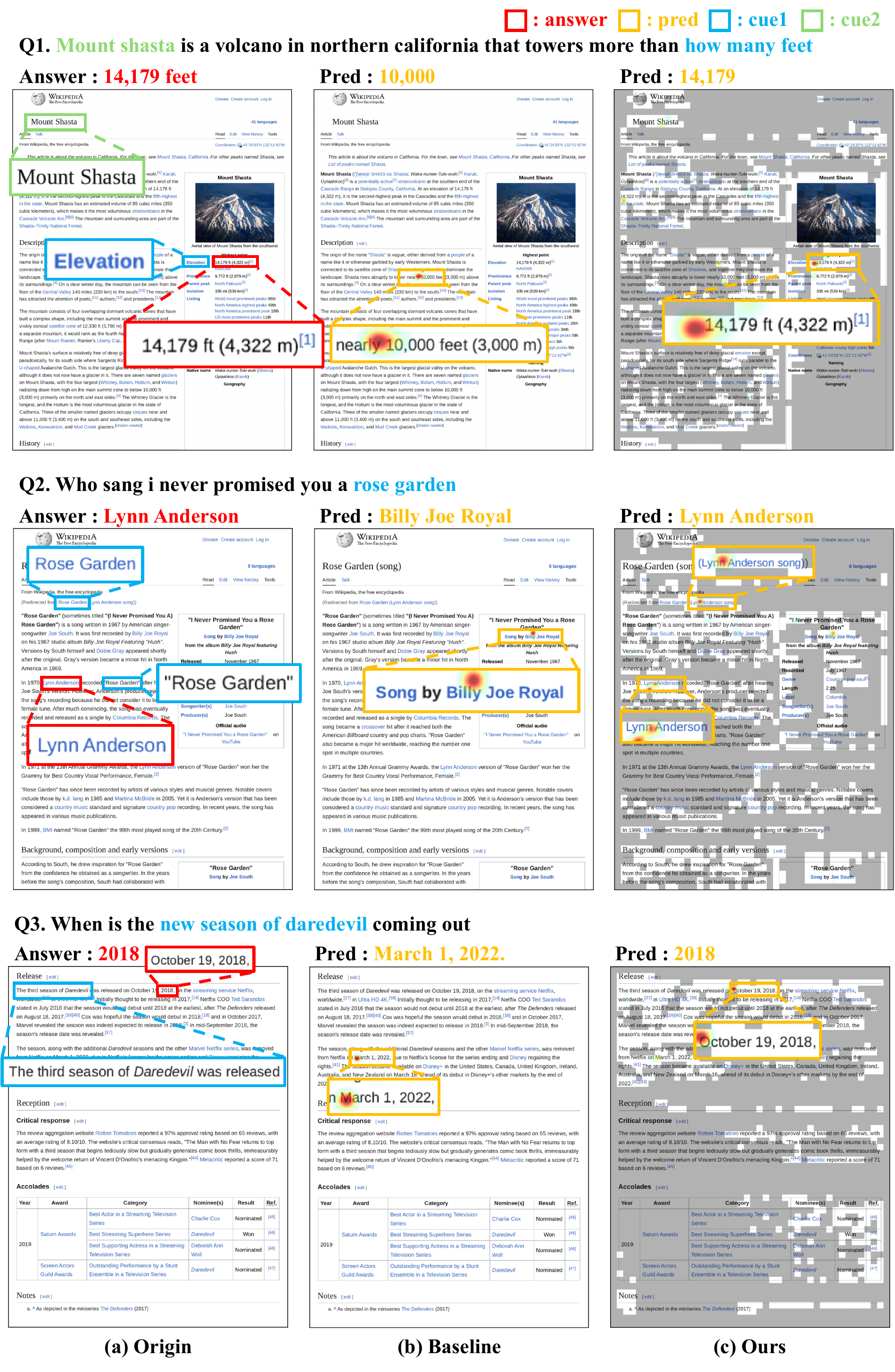}
\end{center}
\caption{ 
\textbf{Qualitative results on irrelevant-region suppression.}
}
\label{fig:fig_11}
\end{figure*}

We qualitatively analyze how removing irrelevant regions affects QA predictions.
Fig.~\ref{fig:fig_11} presents the examples illustrating how removing irrelevant regions improves question-answering performance. 
Given the original document in the left column, the middle and right columns show the attention map referenced by the last token, visualized with and without token pruning, respectively.
The baseline often focuses on noisy or irrelevant areas of the document, leading to incorrect predictions. 
In contrast, our method suppresses unrelated regions and emphasizes areas aligned with the question, enabling the model to produce the correct answer.

\begin{figure*}[ht!]
\begin{center}
\includegraphics[width=0.80\textwidth]{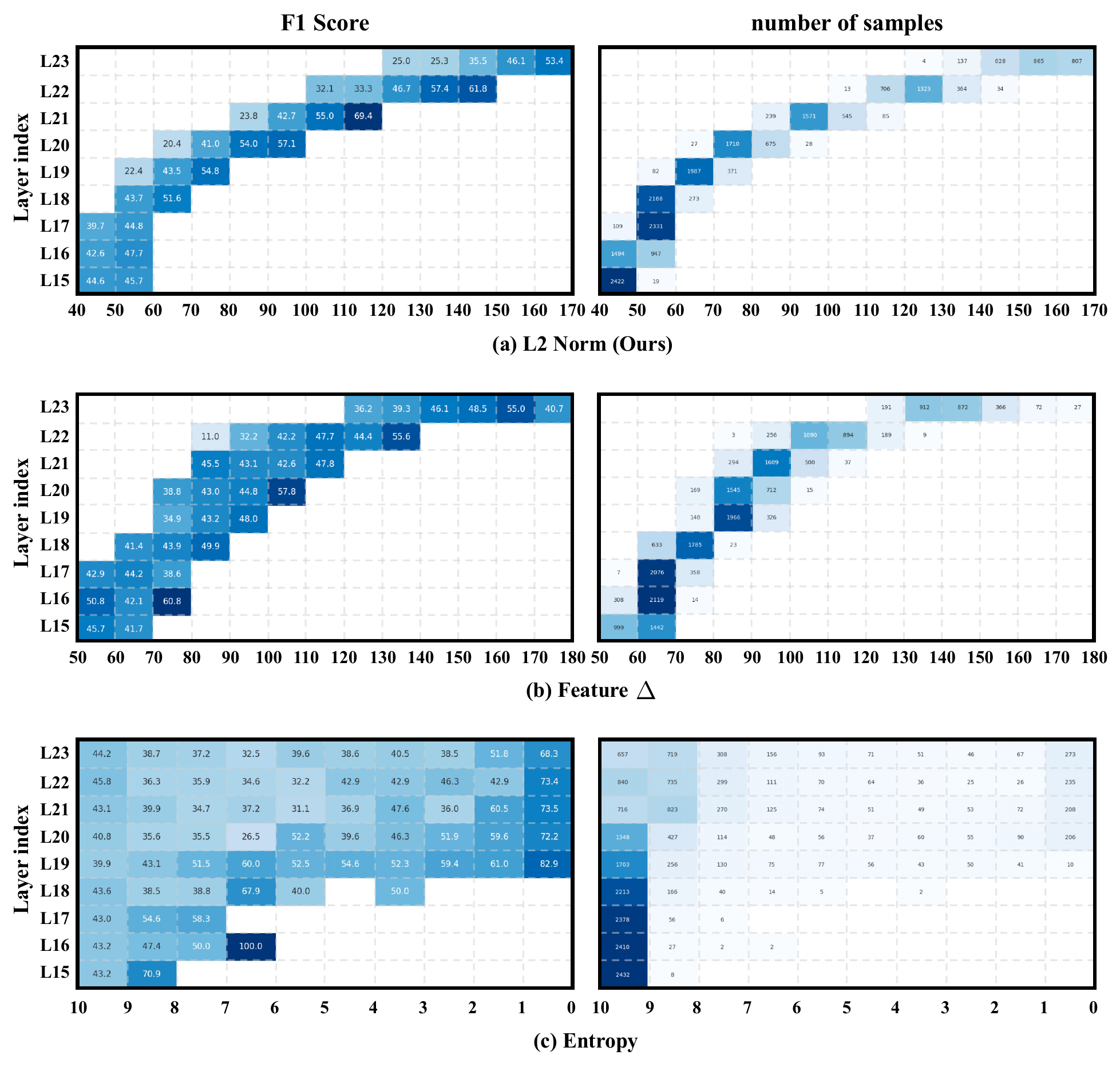}
\end{center}
\vspace{-0.5cm}
\caption{ 
\textbf{Performance and number of samples by layers and multiple criteria.}
}
\label{fig:fig_12}
\end{figure*}

\begin{figure*}[ht!]
\begin{center}
\vspace{-0.5cm}
\includegraphics[width=0.80\textwidth]{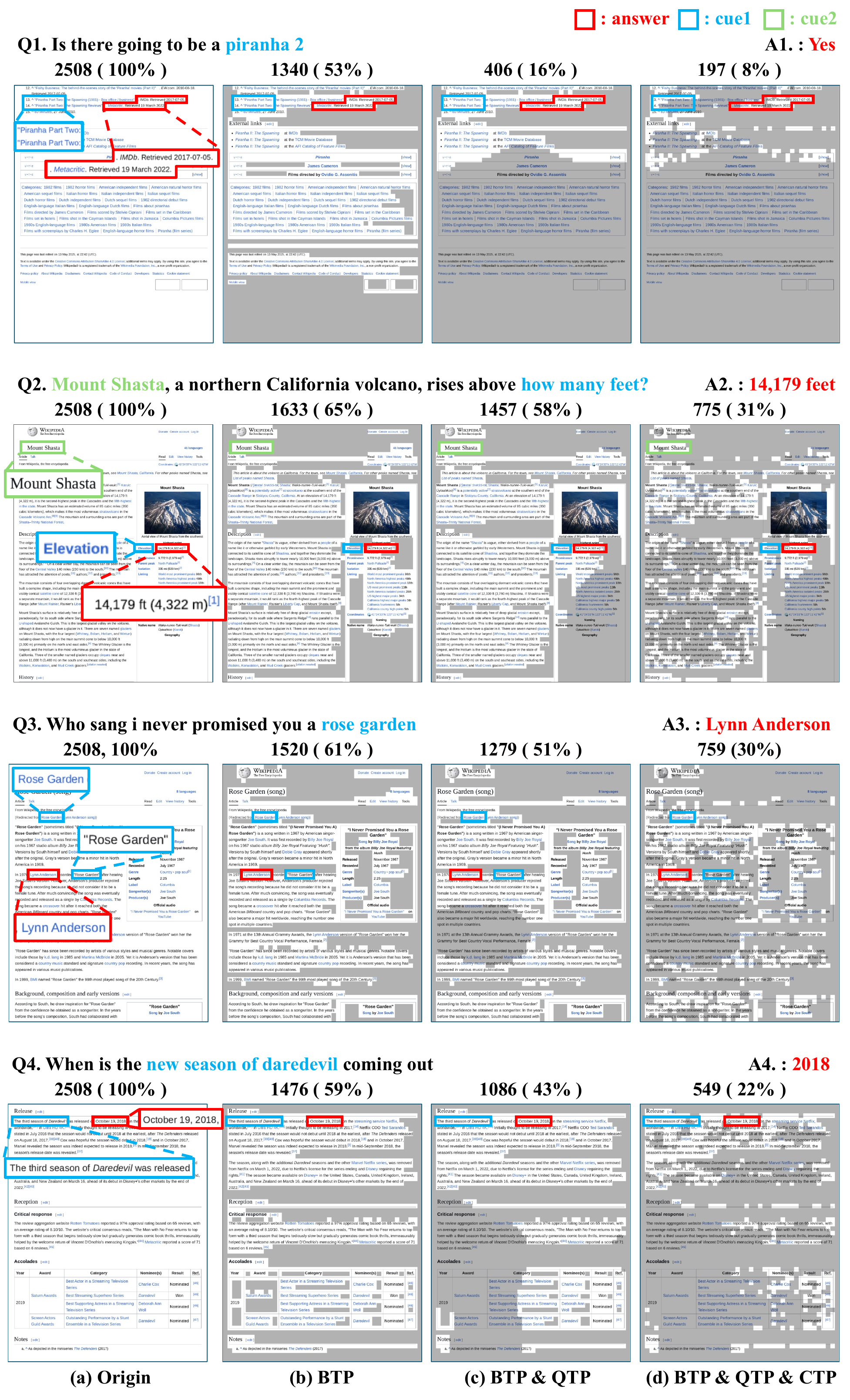}
\end{center}
\vspace{-0.5cm}
\caption{ 
\textbf{Additional Qualitative results of DocPrune.}
}
\vspace{-0.5cm}
\label{fig:fig_13}

\end{figure*}

\section{Comparison of comprehension metrics} 
Fig.~\ref{fig:fig_12} analyzes the relationship between F1 scores and the comprehension criteria in Tab.4 of the main paper.
For each metric, the left heatmap shows the average F1 within each value interval, while the right heatmap shows the number of samples in the corresponding interval.
(a) \textbf{L2 Norm} shows a clear and stable relationship with accuracy. Samples with higher norm values consistently achieve higher F1 scores, and samples that reach such values at earlier layers also tend to perform better. This is consistent with our hypothesis that easier samples attain a confident representation more quickly, so high L2 norms emerging in shallow layers indicate cases that the model can answer correctly with fewer computation steps.
Next, (b) \textbf{Feature $\Delta$} shows a weaker relationship, with many intervals exhibiting high metric values but low accuracy. For example, in Layer 21, the 80-90 interval achieves an average F1 of 45.5, which is higher than the 90-100 and 100-110 intervals (43.1 and 42.6), despite having a lower metric value.
The last metric, (c) \textbf{Entropy}, is the least informative, showing the weakest correlation with the F1 score.
Although lower entropy values often correspond to higher F1, this relationship is inconsistent across intervals, and most samples cluster in the 9–10 range across layers, making entropy a poor discriminator of comprehension.
Overall, the L2 norm provides the most reliable indicator of layer-wise comprehension.

\section{Qualitative results of \textsc{DocPrune}} 
Fig.~\ref{fig:fig_13} presents additional qualitative results.
The figure illustrates how the number and proportion of remaining tokens change as BTP, QTP, and CTP are sequentially applied.
These examples show that \textsc{DocPrune} adaptively adjusts the pruning ratio at each stage for each document–question pair, removing substantial redundancy while preserving the semantic evidence necessary for accurate answers.

\section{Resource Availability}
To support the public release and ensure reproducibility, we provide the official links to the models and datasets utilized in our experiments:
\begin{itemize}
\item \textbf{Qwen2-VL-7B}: \url{https://huggingface.co/Qwen/Qwen2-VL-7B-Instruct}
\item \textbf{Qwen2.5-VL-7B}: \url{https://huggingface.co/Qwen/Qwen2.5-VL-7B-Instruct}
\item \textbf{ColPali-v1}: \url{https://huggingface.co/vidore/colpali-v1}
\item \textbf{VDocRetriever}: \url{https://huggingface.co/NTT-hil-insight/VDocRetriever-Phi3-vision}
\item \textbf{M3DocVQA}: \url{https://huggingface.co/datasets/m3docrag/m3docvqa}

\item \textbf{MMLongBench-Doc}: \url{https://huggingface.co/datasets/m3docrag/mmlongbench-doc}
\item \textbf{OpenDocVQA}: \url{https://huggingface.co/datasets/NTT-hil-insight/OpenDocVQA}
\end{itemize}

\clearpage

\end{document}